\documentclass[10pt,journal,compsoc]{IEEEtran}

\usepackage{color}
\usepackage{float}
\usepackage{graphicx}
\usepackage{multirow}
\usepackage{subcaption}
\usepackage{url}

\ifCLASSOPTIONcompsoc
  \usepackage[nocompress]{cite}
\else
  \usepackage{cite}
\fi

\ifCLASSINFOpdf
\else
\fi
\begin{document}

\title{A hybrid CNN-RNN approach for survival analysis in a Lung Cancer Screening study}
%
%
%
%

\author{Yaozhi~Lu,
        Shahab~Aslani,
        An Zhao,
        Ahmed Shahin,
        David Barber,
        Mark~Emberton,
        Daniel~C.~Alexander, 
        and Joseph~Jacob
\IEEEcompsocitemizethanks{\IEEEcompsocthanksitem Corresponding author: Y. Lu, E-mail: yz.lu@ucl.ac.uk\\
\IEEEcompsocthanksitem Y. Lu, S. Aslani, A. Zhao, A. Shahin, D. Alexander, and J. Jacob are affiliated with the Centre for Medical Image Computing, University College London, UK.
\IEEEcompsocthanksitem Y. Lu, A. Zhao, A. Shahin, D. Barber, and D. Alexander are affiliated with the Department of Computer Science, University College London, UK.
\IEEEcompsocthanksitem S. Aslani and J. Jacob are affiliated with the Department of Respiratory Medicine, University College London, UK.
\IEEEcompsocthanksitem D. Barber is affiliated with the Centre for Artificial Intelligence, University College London, UK.
\IEEEcompsocthanksitem M. Emberton is affiliated with the Division of Surgery and Interventional Science, University College London, UK.
}
\thanks{Manuscript received March X, 2023; revised XX XX, 2023.}}

%
%

\markboth{Preprint}%
{Shell \MakeLowercase{\textit{Lu et al.}}: Bare Demo of IEEEtran.cls}
%



\IEEEtitleabstractindextext{%
\begin{abstract}
In this study, we present a hybrid CNN-RNN approach to investigate long-term survival of subjects in a lung cancer screening study. Subjects who died of cardiovascular and respiratory causes were identified whereby the CNN model was used to capture imaging features in the CT scans and the RNN model was used to investigate time series and thus global information. The models were trained on subjects who underwent cardiovascular and respiratory deaths and a control cohort matched to participant age, gender, and smoking history. The combined model can achieve an AUC of 0.76 which outperforms humans at cardiovascular mortality prediction. The corresponding F1 and Matthews Correlation Coefficient are 0.63 and 0.42 respectively. The generalisability of the model is further validated on an 'external' cohort. The same models were applied to survival analysis with the Cox Proportional Hazard model. It was demonstrated that incorporating the follow-up history can lead to improvement in survival prediction. The Cox neural network can achieve an IPCW C-index of 0.75 on the internal dataset and 0.69 on an external dataset. Delineating imaging features associated with long-term survival can help focus preventative interventions appropriately, particularly for under-recognised pathologies thereby potentially reducing patient morbidity.
\end{abstract}

}

\maketitle

\IEEEdisplaynontitleabstractindextext

%
\IEEEpeerreviewmaketitle

\IEEEraisesectionheading{\section{Introduction}\label{sec:introduction}}

\subsection{Overview}

\IEEEPARstart{C}{ardiac} and respiratory illnesses are the leading causes of mortality globally \cite{roth2018global, vos2020global}, especially amongst older age groups. Compounded with the ageing global population \cite{united2022world}, this translates to an increased number of patients with multimorbid conditions and causes spiralling pressure on healthcare services. Responding to such growing healthcare needs requires cost-effective approaches to early disease detection. Early detection allows timely intervention before diseases become irreversible. 

Various large-scale Lung Cancer Screening (LCS) studies, such as NELSON (Dutch-Belgian)\cite{de2020reduced}, SUMMIT (UK)\cite{horst2020delivering}, the National Lung Screening Trial (NLST, US)\cite{nlst2010studydesign, national2011reduced}, have been set up to improve the detection of early lung cancer in high-risk populations. Screening involves annual computed tomography (CT) imaging and various clinical measurements (e.g. lung spirometry). LCS with CT imaging has been demonstrated to reduce lung cancer mortality in the NLST effectively \cite{nlst2010studydesign,national2011reduced,aberle2013results} and NELSON studies\cite{de2020reduced}. 

A critical and currently underutilised benefit of imaging acquired as part of LCS is the potential to detect underdiagnosed illnesses in  LCS populations. For example, Cardiovascular Disease (CVD) shares the same risk factors as lung cancer. Accordingly, cardiovascular risk detection could be improved by a detailed interrogation of LCS imaging. Similarly, detecting early respiratory pathologies that influence survival could also be improved. Currently, lung cancer detection is typically the radiologist's overarching aim, but improving the detection of cardio-respiratory morbidity could enhance LCS's cost-effectiveness and overall health benefits.

Rather than analysing images in isolation, this study proposes a hybrid CNN-RNN approach to fully utilise both the imaging and temporal information to predict long-term survival outcomes in a LCS cohort. In turn, this may allow early interventions that may prevent or delay the occurrence of adverse cardio-respiratory health events, thereby prolonging a patient's life expectancy. 

\subsection{Literature Review}

Lung cancer and Cardiovascular Diseases (CVD) share several similar risk factors, including smoking (both active and passive) and exposure to fine particulates from air pollution \cite{centers2010tobacco}. Though the pathophysiological mechanisms differ, it has been shown that smoking leads to increased mortality risk from both lung cancer and CVD \cite{pope2011lung, centers2010tobacco}. It is, therefore, logical that patient cohorts enriched with heavy smokers, such as LCS studies \cite{nlst2010studydesign, de2020reduced, horst2020delivering} can be used to develop prediction models of CVD-related mortality. 

Recently, attempts have been made to predict CVD-related mortality in the NLST cohort \cite{van2019direct, guo2019knowledge}. As demonstrated in van Velzen et al. 2019\cite{van2019direct}, a Convolutional Autoencoder (CAE) was trained to derive abstract image features. Then, the extracted features were fed into three separate classifiers to predict CVD-related mortality. The CAE encoded the automatically extracted 3D LDCT volume around the heart and exported the image features to the subsequent classifiers. The support vector machine classifier achieved performance in the Area under ROC curve (AUC) of 0.72. Though the study recognised the value of using clinical information for prediction, including handcrafted variables such as the Coronary Artery Calcium (CAC) score \cite{jacobs2012coronary, chiles2015association}, which is a known predictor of CVD, such information was not utilised in making predictions. Instead, the study demonstrated that it is possible to predict CVD-related mortality from LDCT scans alone. 

Predicting CVD-related mortality was further improved by Guo et al. 2020 \cite{guo2019knowledge}. A multimodal approach was adopted in this study, where models incorporated both LDCT imaging information and handcrafted features to make mortality predictions. When the contributions of the imaging features and clinical data were optimised, this approach improved the AUC performance to 0.82. Both methods show improvement over human performance in this regard. In fact, as reported by Guo et al., visual inspection of the coronary artery calcium measured by a radiologist could only achieve performance with an AUC of 0.64.   

Instead of examining medical images in isolation without their global context, a few recent studies considered adopting a hybrid model architecture to analyse the imaging and temporal features in a patient's data. Thus, hybrid models, which have a CNN portion to investigate the medical imaging features while a RNN portion to keep track of the temporal information,  had been adopted to examine retinal video \cite{gheisari2021combined}, spectrogram \cite{petmezas2022automated}, chest radiograph \cite{santeramo2018longitudinal}, and computed tomography \cite{gao2019distanced} etc. Thus, global temporal information embedded in the follow-up scans can be fully utilised. In the above-mentioned cases, the hybrid models outperformed their CNN-only counterparts, which naturally limited the contextual information from the patient's follow-up history. Thus, the findings illustrate the superior performance of using time-series data over a single snapshot in making the diagnosis.

Given the unavoidable heterogeneous follow-up times in clinical practices, various studies \cite{baytas2017patient, santeramo2018longitudinal, gao2019distanced} had attempted to incorporate the interval variation in their analysis. Instead of adopting the implicit homogeneous interval assumption in the LSTM model, various approaches \cite{baytas2017patient, santeramo2018longitudinal, gao2019distanced} of adding time-weighted terms to accentuate the temporal irregularities are proposed. For example, Santeramo et al. 2018 \cite{santeramo2018longitudinal} proposed additional interval time-related weights to the LSTM gates calculation, while Gao et al. 2019 \cite{gao2019distanced} proposed using the Temporal Emphasis Modules (TEM) to emphasise the more recent scans. 

An alternative and more informative approach to evaluating survival would be the Cox Proportional Hazard (PH) model \cite{cox1972regression} which examines the effect of covariates on the patient's time to death. Katzman et al. 2018\cite{katzman2018deepsurv} proposed, DeepSurv, a Cox Proportional Hazard neural network to examine clinical data to help provide a personalised approach to predict a patient's survival time. A few studies \cite{mobadersany2018predicting, shahin2022survival} have applied a similar approach to survival analysis using medical images. In particular,  Shahin et al. 2022\cite{shahin2022survival} performed survival analysis in an Idiopathic Pulmonary Fibrosis cohort with a modified Residual Network and an equivalent Cox PH loss function.

In our previous work, Lu et al. 2021 \cite{lu2022nlst}, 3D-CNN-based models were deployed to inspect the latest screening CT scan and correlate the imaging features to long-term survival status. A multimodal approach utilising the clinical information was also attempted. However, as discussed in the literature review, such approaches under-utilise the global information in the LCS study and discard an individual's speed of disease progression. Therefore, in this study, we adopted a hybrid CNN-RNN approach to treat the patients' follow-up scans as time series data and combined the local and global information for diagnosis. As can be demonstrated, the hybrid approach can enhance model performance and, in particular, improve their generalisability. When applied to the prediction of time to death, the application of such a method may help direct timely early medical or therapeutic interventions that could prevent or lessen adverse cardio-respiratory events, thus improving morbidity and mortality.

The key contributions of the study are as follows:
\begin{itemize}
    \item A hybrid 4D CRNN approach to predict long-term survival using the patients' longitudinal LCS imaging history.
    \item Predict a patient's risk, thereby suggesting timely personalised interventions to improve patient morbidity and mortality.  
\end{itemize}

\section{Methodology}
\subsection{Dataset selection and split by screening centres}

The National Lung Screening Trial was a lung cancer screening study conducted with 33 US screening centres from 2002 to 2007 \cite{nlst2010studydesign,national2011reduced,aberle2013results}. 53,454 heavy smokers aged 55-74 at high risk for developing lung cancer were recruited. For our study, we were provided with a subset of 15,000 patients' imaging data comprising three annual screening CT attendances, denoted as T0-T2. The participants' survival status and the cause of death were ascertained through the evaluation of death certificate ICD-10 (International Classification of Diseases, 10th edition) codes. In the LDCT branch, the leading cause of death (as of the end of 2009) was cardiovascular disease (26.1\%), followed by lung cancer (22.9\%) and then other types of cancer (22.3\%) \cite{national2011reduced}. 

In our study, the participant's survival status in 2015 was chosen as the ground-truth label, and the follow-up times were adjusted with respect to the latest scan (i.e. T2). The targeted causes of death were cardiac\footnote{ICD-10 codes: I10-I52} and respiratory diseases\footnote{ICD-10 codes: J00-J99}. Only cases with all three screening CTs (i.e. T0-2) were included to avoid bias introduced by patients who left the trial early for unknown reasons. Only CT scan thickness in the axial plane of less than 2.5mm was analysed. A radiologist reviewed the corresponding CT images to exclude cases with imaging artefacts and anatomical biases (i.e. severe forms of thoracic spinal scoliosis). The patients with cardiac and respiratory deaths were age, gender, and smoking history matched in a 1:2 ratio with a control population of survivors.

2154 patients from 32 NLST screening centres meet the eligibility criteria mentioned above. CT scans from 26 of the 32 centres were used as the 'internal' training-validating-testing dataset (n=1,869). Scans from the remaining 6 centres were kept aside as the 'external' dataset (n=285) for later evaluation of the generalisability of the models on heterogeneous imaging protocols and equipment (i.e. out-of-distribution data). The composition of the datasets is tabulated in Table \ref{TAB_Datasets_specifics}. The current combination of the centres in the subsets allows the maximum number of scans to be used.

\begin{table}[ht]
\centering
\begin{tabular}{ll|c|cc|c}
\hline
\multicolumn{2}{c|}{Dataset} & Survivor & \multicolumn{2}{c|}{Non-survivor} & Total \\
             &               &          & Cardiac       & Respiratory       &       \\ \hline
Internal     & CV fold 1     & 208      & 47            & 57                & 312   \\
             & CV fold 2     & 208      & 56            & 48                & 312   \\
             & CV fold 3     & 208      & 59            & 45                & 312   \\
             & CV fold 4     & 208      & 62            & 42                & 312   \\
             & CV fold 5     & 208      & 50            & 54                & 312   \\
             & Test    & 206      & 53            & 50                & 309   \\ \hline
External     & Test    & 190      & 47            & 48                & 285   \\ \hline
Total        &               & 1436     & 374           & 344               & 2154  \\ \hline
\end{tabular}
\caption{Composition of the datasets.}
\label{TAB_Datasets_specifics}
\end{table}

\begin{figure}[h]
  \centering
  \includegraphics[width=1.0\linewidth,trim={0cm 0cm 0cm 0cm},clip]{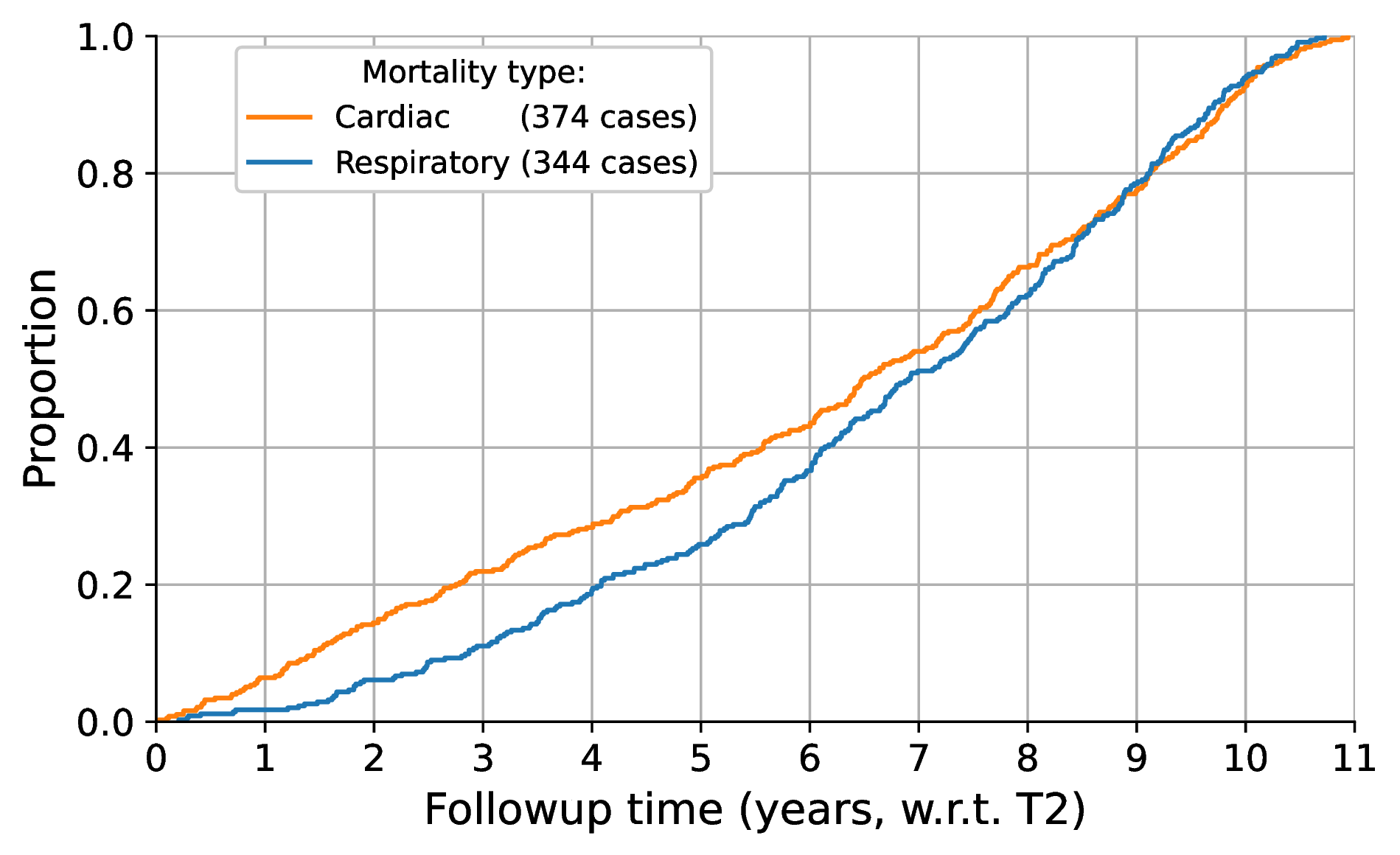}
  \caption{Distribution of the non-survivors' follow-up time.}
  \label{FIG_Nobiko_CNR_dist}
\end{figure}

As shown in Table \ref{TAB_Datasets_specifics}, after a random split, 1560 cases from the internal dataset (n=1,869) were used for a 5-fold cross-validation while the remaining 309 cases were used as an internal test set to assess model performance during training. The composition of the subsets was kept the same throughout the study. The trained models were evaluated on 285 patients' scans in the external test. Additionally, the relative proportions of the two causes of mortality, which are also presented in Table \ref{TAB_Datasets_specifics}, were maintained across the internal and external datasets. The time to death of the non-survivors from the date of the last screening point is illustrated in Figure \ref{FIG_Nobiko_CNR_dist}.

\subsection{Lung CT-volume preprocessing}

\begin{figure}[!t]
  \centering    
  \begin{subfigure}[b]{\linewidth}
    \centering\includegraphics[width=0.56\linewidth,trim={0.0cm 0.0cm 0.0cm 0.0cm},clip]{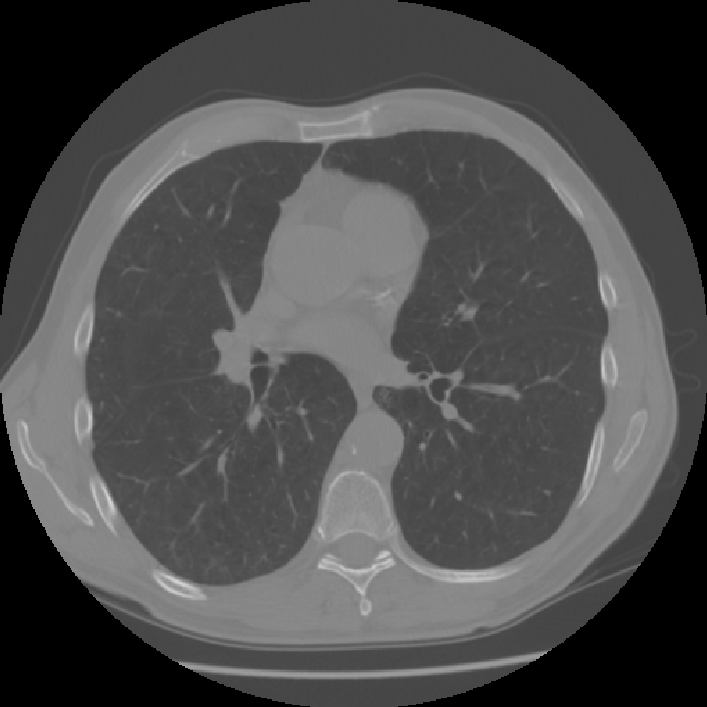}
    \caption{Raw axial slice at 512 by 512 voxels per slice.}    
  \end{subfigure}
  \vspace{-0.2cm}

  \begin{subfigure}[b]{\linewidth}
    \centering\includegraphics[width=0.56\linewidth,trim={0.0cm 0.0cm 0.0cm 0.0cm},clip]{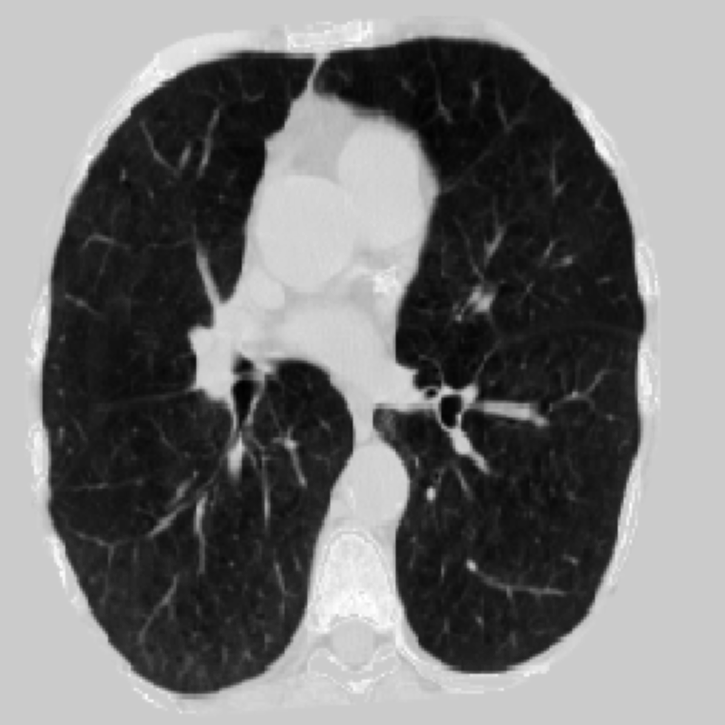}
    \caption{Slice processed and scaled to 256 by 256 voxels per slice.}    
  \end{subfigure}
  
  \caption{CT volume pre-processing.}
  \label{fig_preprocessing}
\end{figure}

Similar to the previous study \cite{lu2022nlst}, the primary imaging processing pipeline consisted of a modified pre-processing approach from Liao et al. \cite{liao2019evaluate}. The CT image was filtered with a Gaussian filter for each axial slice, and a -600 Hounsfiled Unit (HU) threshold was applied to create the binarised slice. The absolute size and eccentricity of connected components were then used to filter out small components and imaging noise. The resulting 3D volumes were then filtered by their size (0.68 - 7.5 L) and distance to the centre of the scan. The results were joined to create the approximate lung mask. Morphological transformations, i.e. erosion, dilation, and convex-hull calculation, were performed to further separate the result into the left and right lung masks. An additional convex-hull analysis was performed on the joint lung masks to include the cardiac region in the final results. The Hounsfield Unit (HU) range was clipped to an interval between -1200 HU to 600 HU. The range was linearly normalised to between 0 and 255. The regions outside the masks corresponding to the surrounding tissue were filled with an average value of 170. The pre-processing pipeline was applied over all axial layers to extract the thorax region in 3D. An example case is illustrated in Figure \ref{fig_preprocessing} where Figure \ref{fig_preprocessing}(a) is a CT scan in DICOM format while Figure \ref{fig_preprocessing}(b) is the pre-processed input to the neural networks.

\begin{figure}[!t]
  \centering
  \begin{subfigure}[b]{\linewidth}
    \centering\includegraphics[width=0.56\linewidth,trim={0.0cm 0.0cm 0.0cm 0.0cm},clip]{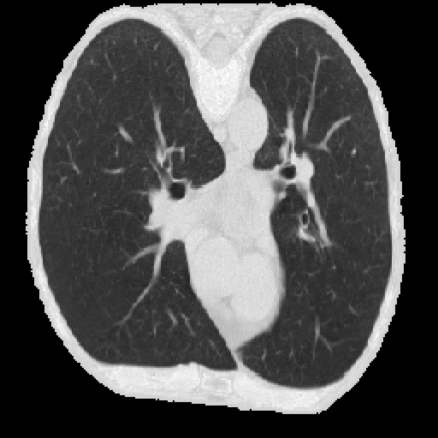}
    \caption{T0 scan.}    
  \end{subfigure} 
  \vspace{-0.2cm}

  \begin{subfigure}[b]{\linewidth}
    \centering\includegraphics[width=0.56\linewidth,trim={0.0cm 0.0cm 0.0cm 0.0cm},clip]{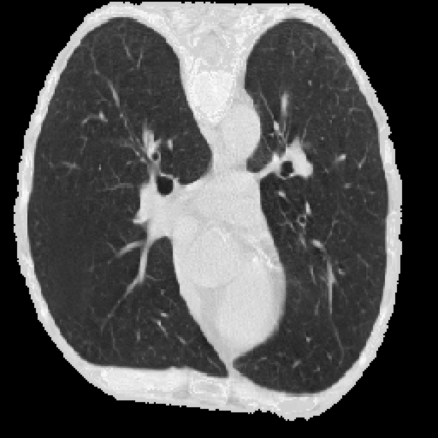}
    \caption{T2 scan.}   
  \end{subfigure}
  \vspace{-0.2cm}
    
  \begin{subfigure}[b]{\linewidth}
    \centering\includegraphics[width=0.56\linewidth,trim={0.0cm 0.0cm 0.0cm 0.0cm},clip]{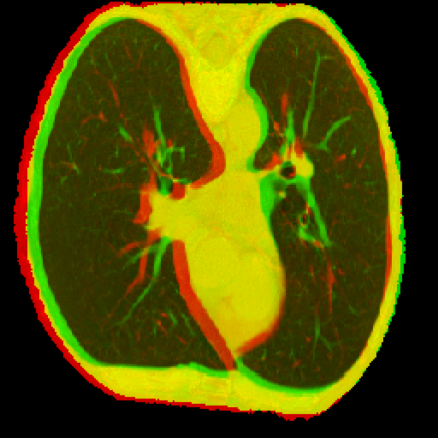}
    \caption{Overlay of T2 and pre-registration T0 scans.}    
  \end{subfigure}
  \vspace{-0.2cm}

  \begin{subfigure}[b]{\linewidth}
    \centering\includegraphics[width=0.56\linewidth,trim={0.0cm 0.0cm 0.0cm 0.0cm},clip]{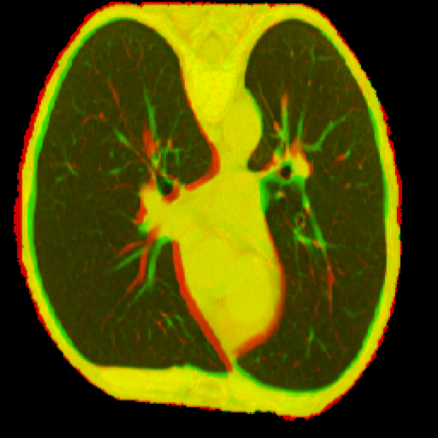}
    \caption{Overlay of T2 and registered T0 scans.}    
  \end{subfigure}
  
  \caption{3D affine registration demonstrated on 2D slices.}
  \label{fig_AFR3D}
\end{figure}

The imaging pre-processing pipeline described above does not explicitly perform image registration. However, it does centre the image based on the joint mask. To assess whether image registration can lead to performance improvement, particularly for the time-series models, 3D affine registration was attempted for one of the sub-studies. In particular, the scans from the two earlier screening points, T0 and T1, were registered to the most recent T2 scan through the DIPY\footnote{The total processing time for 2154 patients, when fully parallelised on an Intel i7-10700K CPU, is approximately 20 hours while running on a SSD.} (v1.5.0) library \cite{garyfallidis2014dipy}. The example results of the 3D affine registration are illustrated in Figure \ref{fig_AFR3D}. Comparing Figure \ref{fig_AFR3D}(c) to Figure \ref{fig_AFR3D}(d), it can be seen that the additional processing has minimised the difference in position and orientation between the T0 scan to the target T2 scan.

\subsection{Models}
\subsubsection{CNN models}

The deep-learning-based models in this study were based on a 3D implementation of the ResNet \cite{he2016deep}\cite{chen2019med3d}. A 10-layer implementation of the 3D ResNet was used as the backbone for both the CNN models and the subsequent CRNN models in this study. The choice of the CNN network depth reflects the trade-off between performance and model size. The pre-trained weights from Chen et al.\cite{chen2019med3d}, initially optimised for medical imaging (CT images) segmentation tasks,  were used as initial weights during training. The output from the 3D ResNet backbone was converted to a 1D tensor after passing through a global adaptive average pooling layer. It was then passed through two subsequent fully-connected layers (including dropout with $p=0.5$ and ReLU activation) with 512 and 32 neurons, respectively.

The CNN models were trained with the latest scan, i.e. NLST T2 scan. The 3D CT volumes were interpolated into 256 by 256 by 128 tensors and fed into the networks. The 2015 survival outcome was used as the ground truth label for the classification task between survivors and non-survivors. 

\subsubsection{CRNN models}

\begin{figure}[h]
  \centering
  \includegraphics[width=1.0\linewidth,trim={0cm 0cm 0cm 0cm},clip]{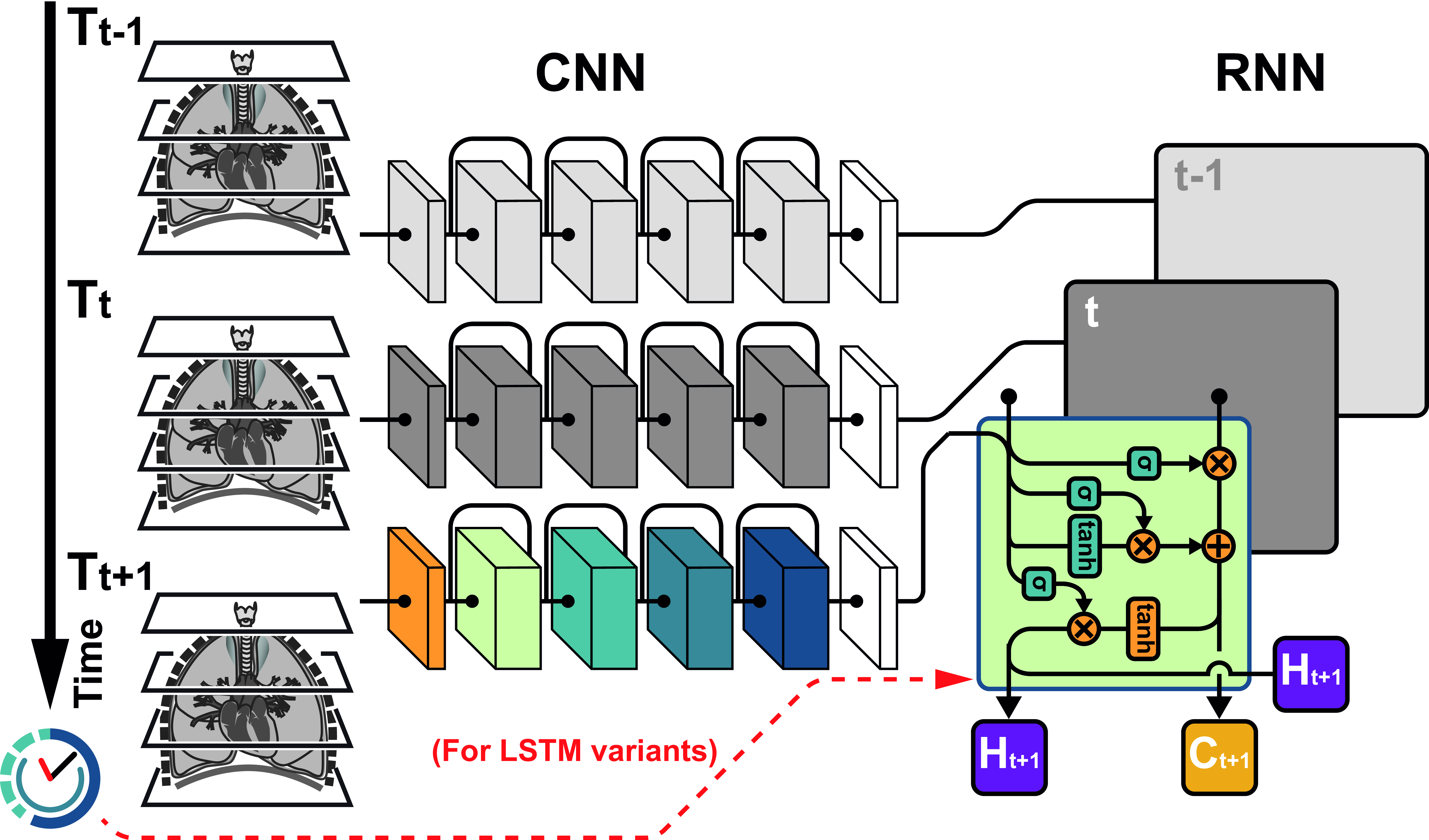}
  \caption{Schematic diagram of the CRNN model.}
  \label{FIG_Nobiko_models}
\end{figure}

The two sides of the CRNN model, as illustrated in Figure \ref{FIG_Nobiko_models}, were trained independently from each other. The model parameters from the trained CNN model were inherited by the CNN side of the model and were not optimised during the CRNN training. In contrast, the RNN (i.e. LSTM) side, which feeds in the spatial /imaging feature from the CNN side, was optimised during CRNN training. In other words, the CNN backbone was used to capture the local imaging features from each CT scan, while the LSTM model was utilised to capture the global time-series information embedded in the patient's entire follow-up history. The output from the LSTM model was passed through one fully-connected layer (including dropout with $p=0.3$ and ReLU activation). The number of neurons in the fully-connect layer matched the LSTM's hidden state (hx) size, a tunable hyperparameter. 

For each patient, the interpolated 3D CT volumes were stacked into a 4D tensor, with the temporal dimension (i.e. T0-2) as the additional dimension., All 3 NLST screenings were analysed by the hybrid CRNN model, which can process a variable number of scans in a patient's follow-up history. Thus, in this study, the input to the CRNN network was in the form of 256 by 256 by 128 by three tensors.

It is important to note that the vanilla LSTM implementation implicitly assumes the inputs are equally spaced. Given the inevitable irregular time intervals between screening timepoints that occur in medical studies, two additional LSTM variants, Time-Aware LSTM (TALSTM)\cite{baytas2017patient} and time-modulated LSTM (tLSTM)\cite{santeramo2018longitudinal}, which explicitly take irregular time intervals between inputs into account were tested. To account for the influence of non-uniform time intervals, the former model\cite{baytas2017patient} attempts to separate the memory component in LSTM into its long and short-term parts, with the latter's effect disregarded. In contrast, the time-modulated LSTM variant \cite{baytas2017patient} explicitly introduces temporal weights into the LSTM gates update. For both variants, as illustrated by the dashed lines in Figure \ref{FIG_Nobiko_models}, the temporal information was passed directly into the RNN portion of the model. Pre-processing of the temporal data was implemented in the same manner recommended by the respective studies.

\subsubsection{Cox Proportional Hazard Network}
In addition to performing classification on the long-term survival outcome, a survival outcome prediction based on the Cox Proportional Hazard model\cite{cox1972regression}  was attempted using the same CNN and CRNN networks. In this study, the survival time is right-censored and adjusted with respect to the latest screening point (T2). 

In survival analysis, a hazard function measures the instantaneous risk of event /death ($E=1$) at time $t$ for an individual that has survived beyond time $t$: 

\begin{equation}
    h(t)=\displaystyle{\lim_{\triangle t \to 0}}\frac{p(t\leq T < t+\triangle t | T\geq t)}{\triangle t}
    \label{EQN_coxph_00_h0}
\end{equation}

The Cox PH model attempts to model the hazard function given a patient's baseline data $x$: 

\begin{equation}
    h(t|x)=h_{0}(t)\cdot e^{h(x)}
    \label{EQN_coxph_01_hratio}
\end{equation}

In Eq. \ref{EQN_coxph_01_hratio}, $h_{0}(t)$ is the baseline hazard function and represents the hazard when all the covariates are zero. $h(x)$ in this study is the output of the neural network. To optimise such a network, the negative partial log-likelihood function \cite{katzman2018deepsurv, shahin2022survival} is implemented:

\begin{equation}
    L = - \frac{1}{N_{E=1}}\sum\limits_{i:E_{i}=1}^{}\left(h(x_{i})-\log\sum\limits_{j\in R(T_{i})}^{}e^{h(x_{j})}\right)
    \label{EQN_coxph_02_coxloss}
\end{equation}

where $N_{E=1}$ is the total number of patients with an event and $R(T_{i})$ is the set of patients that have not died before patient $i$ at time $T_{i}$. The loss function $L$ is optimised using stochastic gradient descent.

\subsubsection{Experiments}

A grid search approach was adopted to tune the hyperparameters in the neural network models. The hyperparameter value that produced the best AUC value was chosen.

Similar to the previous work\cite{lu2022nlst}, to counter the class imbalance in the 1:2 matched dataset, a weighted random sampler assigns a sampling probability inversely proportional to the class size was utilised. This approach attempts to create balanced batches during training. Cross entropy loss was adopted as the loss function for the classification model while the Cox Loss in Eq. \ref{EQN_coxph_02_coxloss} was adopted for the regression models. The models were trained with Sharpness Aware Minimisation (SAM)\cite{foret2020sharpness} with Stochastic Gradient Descent (SGD) as the base optimiser. SAM simultaneously minimises the value and sharpness in the training loss landscape. Thus, it aims to optimise for model parameters within the neighbourhood of low loss and hence improves the generalisability of the model. Adaptive implementation of this optimisation algorithm was also tested, yet it did not improve performance.

As a substudy, an additional set of networks was trained to further differentiate the non-survivor classes into the respective mortality groups, i.e. cardiac-related and respiratory-related. This additional insight provides the clinician with further information as to which areas on the CT need specific consideration. A two-step process is proposed to utilise this model in a clinical setting. Firstly, the main model examines a patient's scans to produce a mortality probability. Secondly, the scans are examined by the additional cause-specific model should the previous output exceed a predefined threshold which is 0.5 in this study. The cause-specific model will then attempt to categorise the site of concern further, and the clinician can perform further checks accordingly. Thus, additional models were trained with only the two non-survivor classes shown in Table \ref{TAB_Datasets_specifics} and the composition of each subset was kept the same as the main model. The primary CRNN model's training approach was adopted for this substudy.

The classification models were optimised using an initial learning rate of \textsl{1E-3}, a cyclic learning rate schedule, momentum of \textsl{0.9}, and weight decay (L2 regularisation) of \textsl{5E-4}. The regression models differ regarding the initial learning rate, which was set as \textsl{5E-3}. Each neural network was trained with a batch size of 24 on a single Nvidia A100 GPU (40GB HBM2) on the UCL CS CMIC cluster. The CNN models were trained with approximately 800 epochs, while the CRNN models were trained with approximately 150 epochs. The deep learning models were developed in Python (v3.9.5) and PyTorch (v1.9.1).

\section{Results}
\subsection{Classifying Cardiorespiratory Mortality}

To compare the performance with related studies \cite{van2019direct,guo2019knowledge}, the models were mainly assessed with the Area Under the Curve (AUC) metric. The average (and standard deviation) of the AUC metrics from the 5-fold cross-validation were used to assess the internal performance of the model. As tabulated in Table \ref{TAB_metrics_CORE}, the same was done for the F1 score and Matthews Correlation Coefficient (MCC)\cite{chicco2020advantages}. To evaluate the model's generalisation ability, the average (and standard deviation) inference performance of the five models on the external dataset was also recorded through the three metrics in Table \ref{TAB_metrics_CORE}.

\begin{figure}[!t]
  \centering
  \begin{subfigure}[b]{\linewidth}
    \centering\includegraphics[width=1.0\linewidth,trim={0.2cm 0.2cm 0.2cm 0.2cm},clip]{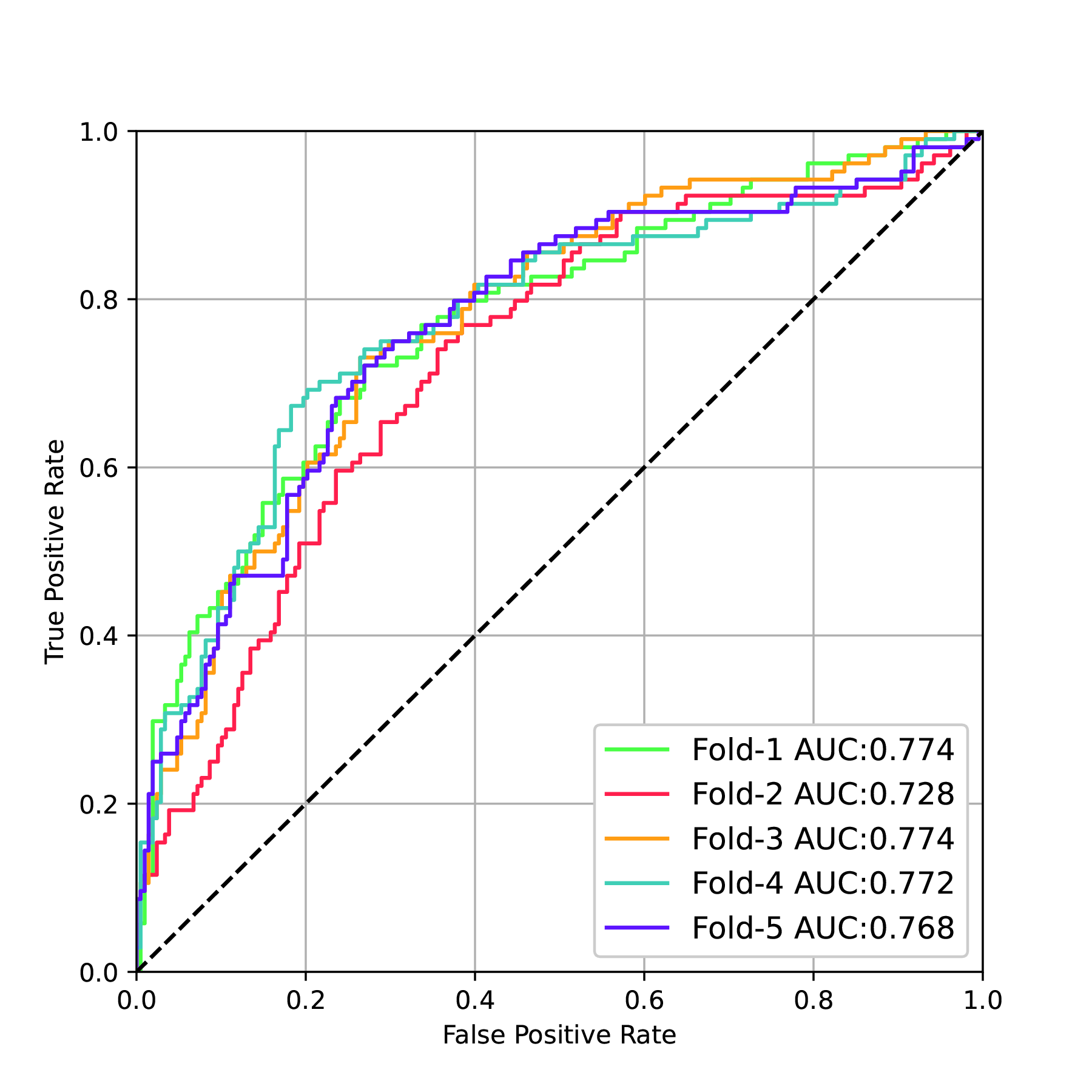}
    \caption{ROC curves from the 5 cross-validation folds (i.e. internal dataset).}   
  \end{subfigure}
  \vspace{-0.5cm}

  \begin{subfigure}[b]{\linewidth}
    \centering\includegraphics[width=1.0\linewidth,trim={0.2cm 0.2cm 0.2cm 0.2cm},clip]{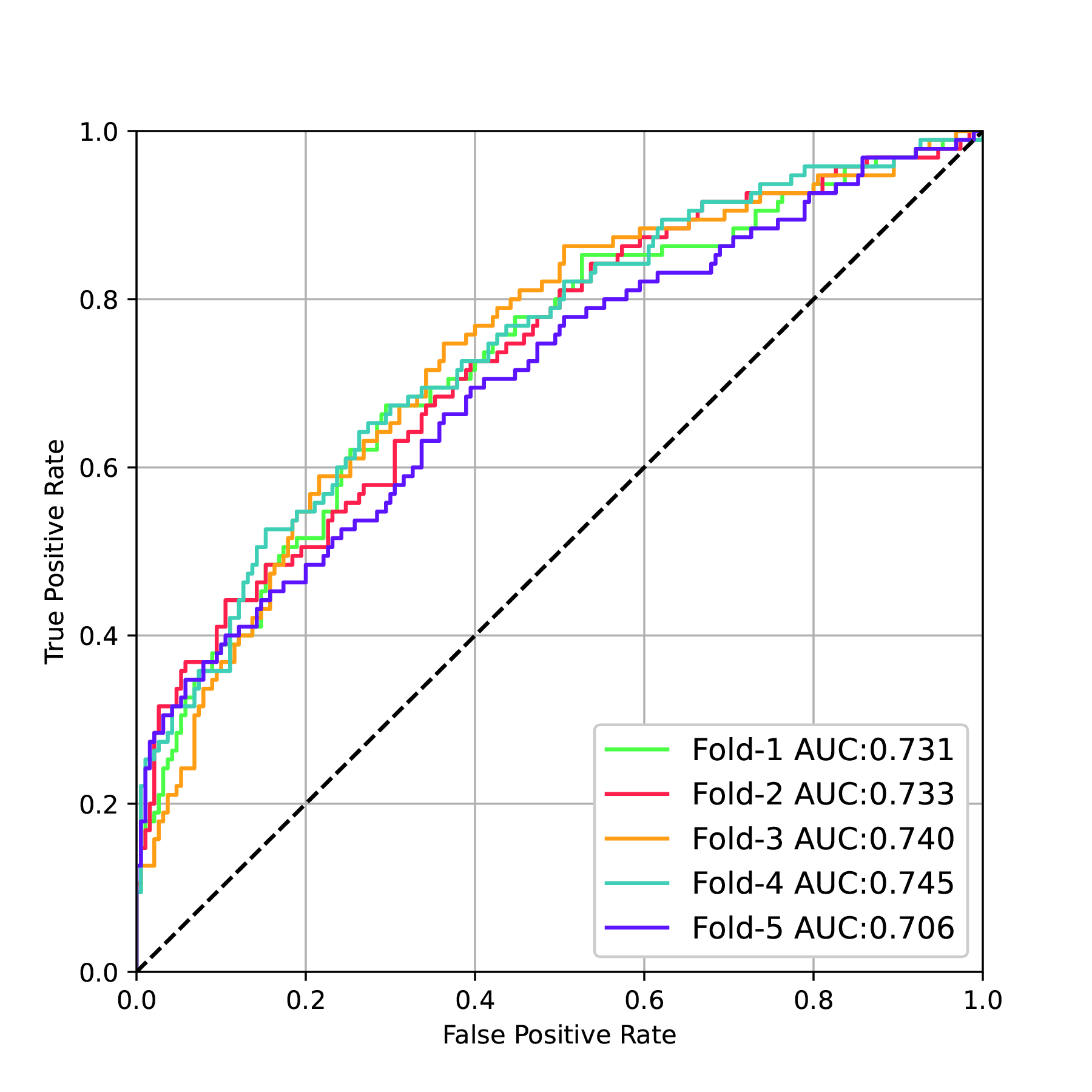}
    \caption{ROC curves from inference on the external dataset.}    
  \end{subfigure}  
  \caption{CRNN model (Study B) performance illustrated by Receiver Operating Characteristic (ROC) curves.}
  \label{fig_roc}
\end{figure}

In the baseline Study A, where the CT scans are analysed in isolation from the rest of the sequence of CT time points, the model reached an average AUC of 0.759, an average F1 of 0.626, and an average MCC of 0.407 on the internal dataset. As expected, the performance in all three metrics deteriorated when evaluated on the external dataset. In comparison, when the global and temporal information from the patient's follow-up was considered in Study B, all three performance metrics on the internal dataset improved over those in Study A. In particular, the Matthews Correlation Coefficient improved by 3.4\% from 0.407 to 0.421. The extent of the performance improvement introduced by the hybrid approach is on par with related studies \cite{gao2019distanced}\cite{santeramo2018longitudinal}. More importantly, the generalisation ability of the model, indicated by performance on the external dataset, improved, as illustrated by the increase in AUC performance from 0.714 to 0.731 in Table \ref{TAB_metrics_CORE}. The corresponding ROC curves for Study B are shown in Figure \ref{fig_roc}.

\begin{figure}[!t]
  \centering
  \begin{subfigure}[b]{\linewidth}
    \centering\includegraphics[width=0.55\linewidth,trim={0.0cm 0.0cm 0.0cm 0.0cm},clip]{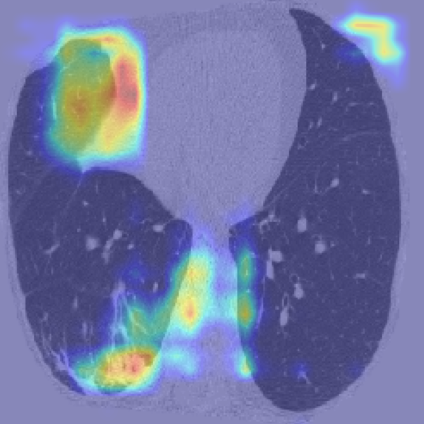}
    \caption{The first screening point (T0).}   
  \end{subfigure}
  \vspace{-0.2cm}

  \begin{subfigure}[b]{\linewidth}
    \centering\includegraphics[width=0.55\linewidth,trim={0.0cm 0.0cm 0.0cm 0.0cm},clip]{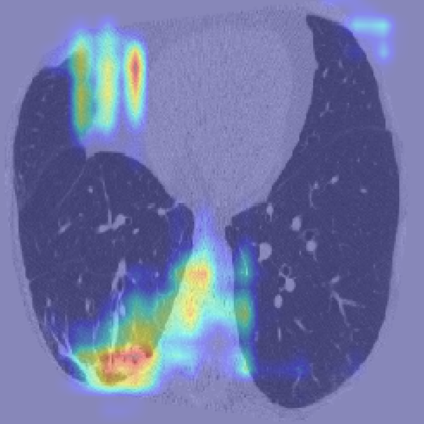}
    \caption{The second screening point (T1).}    
  \end{subfigure} 
  \vspace{-0.2cm}
    
  \begin{subfigure}[b]{\linewidth}
    \centering\includegraphics[width=0.55\linewidth,trim={0.0cm 0.0cm 0.0cm 0.0cm},clip]{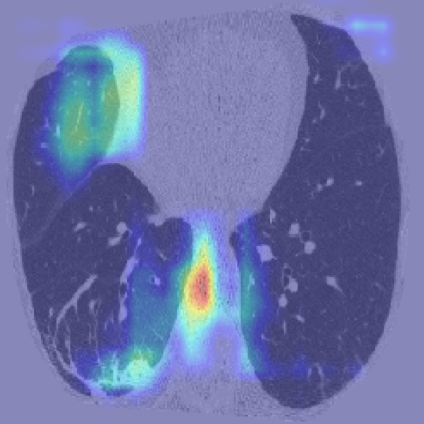}
    \caption{The third screening point (T2).}    
  \end{subfigure}
  
  \caption{GradCAM overlays of the 3 screening points for a COPD non-survivor.}
  \label{fig_saliency_maps}
\end{figure}

Study C, where a 3D registered dataset was used to train the combined CRNN model, showed no improvement in performance on internal and external datasets. It also showed that the additional noise introduced by the affine registration did not outweigh the benefits gained. To further investigate the utility of the 3D registration, GradCAMs (Gradient-weighted Class Activation Map)\cite{selvaraju2017grad} from a Chronic Obstructive Pulmonary Disease (COPD) non-survivor's follow-ups scans are shown. As illustrated by the GradCAM overlays in Figure \ref{fig_saliency_maps}, the model is relatively consistent with lesion position over time, without the explicit registration, thereby negating the need for the additional 3D registration.

\begin{table*}[ht!]
\begin{tabular}{l|cc|c|ccc|ccc}
\hline
Study          & \multicolumn{2}{c|}{Model Component(s)} & LSTM &\multicolumn{3}{c|}{5-Fold CV Performance} & \multicolumn{3}{c}{External Performance} \\
            & CNN           & RNN     & hx   & AUC              & F1              & MCC             & AUC            & F1             & MCC            \\ \hline
A. CNN     & ResNet   & -       & -   & 0.759 (0.019)    & 0.626 (0.026)   & 0.407 (0.047)   & 0.714 (0.011)   & 0.578 (0.010)   & 0.315 (0.021)   \\
B. LSTM       & ResNet   & LSTM    & 32  & 0.763 (0.020)    & 0.629 (0.031)   & 0.421 (0.047)   & 0.731 (0.015)   & 0.571 (0.021)   & 0.319 (0.037)   \\
C. LSTM (3d reg) & ResNet   & LSTM    & 32  & 0.760 (0.017)    & 0.605 (0.048)   & 0.404 (0.042)   & 0.729 (0.014)   & 0.565 (0.030)   & 0.324 (0.034)   \\
D. TALSTM     & ResNet   & TALSTM  & 32  & 0.763 (0.019)    & 0.628 (0.034)   & 0.423 (0.048)   & 0.730 (0.015)   & 0.573 (0.026)   & 0.326 (0.035)   \\
E. tLSTM      & ResNet   & tLSTM & 64  & 0.763 (0.018)    & 0.630 (0.028)   & 0.419 (0.046)   & 0.731 (0.014)   & 0.575 (0.018)   & 0.320 (0.039)   \\ \hline
\end{tabular}
\caption{Performance metrics on long term survival prediction. }
\label{TAB_metrics_CORE}
\end{table*}

In Study D and E, where the irregularity in follow-up intervals are explicitly addressed, the performance is comparable to Study B, which implicitly assumes constant intervals between scans. This implies the follow-up interval variation, which has an inter-quartile range of 40 days, provides limited information to the long-term survival prediction task.

\subsection{Classifying Cause-Specific Cardiorespiratory Mortality}

The CRNN network in Study B was applied to patients who died from cardiac and respiratory deaths, as shown in Table \ref{TAB_metrics_CNR_bands}, where deaths were classified into three bands: 0-3 years from the last CT date (i.e. T2); 0-7 years from the last CT date; within 11 years of the last CT date. The model had initially been trained with labels corresponding to known patient deaths 10 years after the first CT, but had not been provided with information about the exact cause of death (i.e. whether cardiac or respiratory related). Any patient predicted by the ensemble model from Study B to have a less than 0.5 mortality probability was classified as a survivor. Any cases passing this probability threshold were separated into the two causes of death by the CRNN model in Study F. The final prediction from this two-step decoupled process was then evaluated against their ground truth label of dead/alive within the relevant time band. Accordingly, the ground truth survival status varied according to the follow-up time band. 

As shown in Table \ref{TAB_metrics_CNR_bands}, respiratory mortality within 3 years of the CT scan was predicted with good sensitivity (0.737) and specificity (0.765), with cardiac mortality prediction also showing good specificity (0.771) but poor sensitivity (0.476). Respiratory death prediction showed improved specificity and acceptable sensitivity at time intervals increasingly distant from the last available imaging that could be analysed. Cardiac death prediction however showed poor sensitivity and good specificity across all follow-up time bands.

\begin{table*}[]
\centering
\begin{tabular}{l|ccc|ccc|ccc}
\hline
Class           & \multicolumn{3}{c|}{Band a (death \textless 3.0 yrs)}                                         & \multicolumn{3}{c|}{Band b (death \textless 7.0 yrs)}                                          & \multicolumn{3}{c}{Band c (death \textless 11.0 yrs)}                                         \\
                & \multicolumn{1}{c}{Cases} & \multicolumn{1}{c}{Sensitivity} & \multicolumn{1}{c|}{Specificity} & \multicolumn{1}{c}{Cases} & \multicolumn{1}{c}{Sensitivity} & \multicolumn{1}{c|}{Specificity} & \multicolumn{1}{c}{Cases} & \multicolumn{1}{c}{Sensitivity} & \multicolumn{1}{c}{Specificity} \\ \hline
Survivor        & 2034                      & 0.626                           & 0.744                            & 1776                      & 0.688                           & 0.746                            & 1436                      & 0.791                           & 0.719                           \\
NS(Cardiac)     & 82                        & 0.476                           & 0.771                            & 202                       & 0.510                           & 0.806                            & 374                       & 0.513                           & 0.853                           \\
NS(Respiratory) & 38                        & 0.737                           & 0.765                            & 176                       & 0.665                           & 0.808                            & 344                       & 0.634                           & 0.861                           \\ \hline

\end{tabular}
\caption{Performance by time range and cause of mortality (Study F).}
\label{TAB_metrics_CNR_bands}
\end{table*}

\subsection{Predicting Patients' Risk through the Cox model}
Our final analysis considered discriminating the NLST cohort based on their cardio-respiratory mortality risk. The performance of the Cox-regression-based survival analysis was measured across internal and external cohorts and evaluated using the Inverse Probability of Censored Weights (IPCW) Concordance Index \cite{uno2011c} which is modified from the original Concordance Index \cite{harrell1996multivariable}. The tabulated metrics in Table \ref{TAB_metrics_CoxLoss} illustrate that adopting the CRNN model which examines all available follow-up imaging information outperforms the CNN approach which investigates the latest CT scan in isolation. In particular, the IPCW C-index is improved by 4.3\% from 0.719 to 0.750 between the two models. Though less pronounced, an improvement of 2.5\% in IPCW C-index from 0.673 to 0.690 can be observed in the external dataset. This reaffirms our hypothesis that the hybrid CNN-RNN approach improves model performance by capturing additional global time-series CT features that relate to cardio-respiratory disease progression. 

\begin{table}[ht]
\centering
\begin{tabular}{l|l|c|cc}
\hline
CoxPH       & Model         & LSTM  & \multicolumn{2}{c}{IPCW C-Index} \\
Study       &               &  hx   & 5-fold CV     & External         \\ \hline
G. CNN      & ResNet        &  -    & 0.719 (0.040) & 0.673 (0.009)    \\
H. CRNN     & ResNet+LSTM &  32   & 0.750 (0.050) & 0.690 (0.019)    \\ \hline
\end{tabular}
\caption{Regression model performance in terms of average IPCW C-index (and standard deviation).}
\label{TAB_metrics_CoxLoss}
\end{table}

\section{Discussion}

Our study examined the ability of time-series imaging data to predict mortality in subjects participating in a lung cancer screening study. We demonstrate that time series analysis improves the ability of single timepoint data in mortality prediction. We demonstrate the high sensitivity and specificity of our model in estimating the likelihood of respiratory deaths at 3 years, and the high specificity for the detection of cardiac-related deaths across follow-up timepoints. We also demonstrate the improved ability of our time-series model to predict the time to cardiorespiratory death using a modified Cox model when compared to single timepoint data.

\subsection{Survival prediction using both spatial and temporal data}

By comparing the performance between Study A and B, in Table \ref{TAB_metrics_CORE}, it is evident that the inclusion of longitudinal information improves classification performance. When compared to examining CT images in isolation, (Study A where a CNN model is adopted to view the latest CT scan), considering the entire follow-up imaging history of a patient provides a greater global context of the patient's health and/or disease progression. Specifically, the CRNN model improves the internal MCC by 3.4\%  from 0.407 to 0.421 and the external AUC by 2.4\% from 0.714 to 0.731. The extent of the improvement, through the inclusion of temporal information, is consistent with that observed in related studies \cite{gao2019distanced, santeramo2018longitudinal}.

Despite the marginal improvement in external F1 and MCC, the irregular time interval model examined in Study D and E showed similar performance to the model evaluated in Study B. This is in contrast to the performance improvement noted in the related studies \cite{baytas2017patient, gao2019distanced, santeramo2018longitudinal}. It can be argued that this study's classification task, which focuses on long-term survival, is more tolerant to the time interval variation than counterparts, such as the study by Santeramo et al. 2018 \cite{santeramo2018longitudinal}. Yet the most likely cause for the comparable performance between models that did and did not consider time intervals is the relatively constant time interval (inter-quartile range of 40 days between CTs) in the NLST study which was a clinical trial. Models trained to examine irregular time intervals on longitudinal data are far more likely to show utility in real-world data where imaging intervals are more varied than would be seen in protocolised clinical trials.

\subsection{Cause-specific survival outcomes}
Our results demonstrate how differently our model treated respiratory versus cardiac deaths in the screening population. The high sensitivity and specificity for 3-year respiratory mortality suggest the model has potential utility in identifying screened patients who might benefit from targeted respiratory interventions over a meaningful time frame for clinical trials.

The low sensitivity of the model for cardiac death prediction may relate to the inherently unpredictable nature of many acute cardiac events. A central challenge of cardiology lies in risk prediction and prevention for populations at risk of adverse cardiovascular events. Whilst calcification can be identified in coronary arteries on non-contrast-enhanced CTs, calcified plaques are the main surrogates for the presence of non-calcified plaques which are more likely to rupture. Yet coronary calcification and non-calcified coronary damage may be very challenging to identify comprehensively on low-dose, non-gated, non-contrast CT scans. The high specificity for cardiac death prediction identified in our study suggests some events can be predicted using imaging and the results are reassuring in the context of a screening population, where false positive identification of disease should be avoided. Analysis of the time-series CTs in a screening cohort with a model such as ours could help rule-in subjects who would warrant a more definitive evaluation of cardiac disease with for example a contrast-enhanced, gated coronary CT scan.  

The improved performance of the Cox regression model when using time-series data confirms our hypothesis that capturing change in imaging features and therefore disease progression, is important when estimating mortality and has clear advantages over single timepoint data. The standard workflow for a radiologist analysing imaging data relies on the assessment of historic imaging to assess the extent of disease progression. Models utilising the additional information available through repeated imaging therefore should be better at estimating disease progression.

\section{Conclusion and Future Works}
The results shown in our study lead us to the following conclusions and directions for future studies:

\begin{enumerate}
    \item The comparison between the baseline CNN model and the main CRNN model indicates that capturing both local and global information in a patient's medical history can lead to better model performance and improved generalisation ability. Thus, unsurprisingly, medical images should not be viewed in isolation. Instead, longitudinal information should be captured to monitor disease progression. The next step would be deploying the current hybrid model on contemporaneous lung cancer screening studies. The SUMMIT study\cite{horst2020delivering} provides such an opportunity where the ability to predict cardio-respiratory adverse events could be examined prospectively.
    \item It is clear from Study D and E that the limited variation in follow-up intervals across the NLST screening time points did not unduly influence long term survival prediction. It would be important to deploy models that consider irregular time interval acquisitions on non-protocolised real-world data to gauge the influence of irregular scanning intervals on estimations of disease progression. 
    \item An alternative to the proposed CRNN model would be a Vision Transformer (ViT) \cite{dosovitskiy2020image}. A transformer-based architecture was proposed by Sarasua et al. 2021 \cite{sarasua2021transformesh} to model spatio-temporal neuroanatomical changes in a patient's left hippocampus. To account for missing follow-up imaging timepoints, padding was applied to the sequences. We would therefore like to explore a time-dependent variant to our model to account for temporal heterogeneity.
\end{enumerate}

\ifCLASSOPTIONcompsoc
  \section*{Acknowledgments}
\else
  \section*{Acknowledgment}
\fi

This work was supported by the International Alliance for Cancer Early Detection, an alliance between Cancer Research UK [C23017/A27935], Canary Center at Stanford University, the University of Cambridge, OHSU Knight Cancer Institute, University College London, and the University of Manchester. The authors would also like to thank the National Institute of Health Research (NIHR) UCLH  Biomedical Research Centre for the funding. 

The authors thank the National Cancer Institute for access to NCI's data collected by the National Lung Screening Trial (NLST). The statements contained herein are solely those of the authors and do not represent or imply concurrence or endorsement by NCI.

The GPU computing resource of this work is jointly provided by the following funders (and grants):
CRUK International Alliance for Cancer Early Detection (ACED) (\url{C23017/A27935}); CRUK-EPSRC (\url{NS/A000069/1}); MRC-JPND (\url{MR/T046473/1}); Open Source Imaging Consortium; Royal Academy of Engineering (\url{RF\201718\17140}); UKRI /EPSRC /MRC (\url{EP/R006032/1}, \url{EP/R014019/1}, \url{EP/T026693/1}, \url{EP/T029404/1}, \url{EP/V034537/1}, \url{EP/W00805X/1}, \url{MR/S03546X/1}, \url{MR/T019050/1}); Wellcome Trust (\url{209553/Z/17/Z}, \url{221915/Z/20/Z}).

The first author would like to thank Mr T Clark and Mr E Martin from the UCL CS Technical Support Group for their assistance with the CMIC GPU cluster.

\ifCLASSOPTIONcaptionsoff
  \newpage
\fi



%

\bibliographystyle{IEEEtran}
\bibliography{Nobiko_CRNN}

\vfill


\end{document}